%% file: main_iswc.tex
\documentclass[runningheads]{llncs}

\usepackage{amsmath}
\usepackage[T1]{fontenc}
\usepackage{graphicx}
\usepackage{algorithm}
\usepackage{algorithmic}
\usepackage{booktabs}
\usepackage{multirow}
\usepackage{amssymb}
\usepackage{cleveref}
\usepackage{newfloat}
\usepackage{listings}

\usepackage{xcolor}

\newcommand{\model}{DeMix}

\begin{document}

\title{
% HSGenerating: High-quality Sample Generating for Knowledge Graph Embedding \\
% Mix-up-based High-quality Sampling for Knowledge Graph Embedding \\
Negative Sampling with Adaptive Denoising Mixup for Knowledge Graph Embedding
}
% an easy and efficient denoising framework

\author{%
Xiangnan Chen \and
Wen Zhang\and
Zhen Yao\and
Mingyang Chen\and
Siliang Tang\thanks{Corresponding author.}}

\institute{
 Zhejiang University, China \\
\email{\{xnchen2020, zhang.wen,	22151303, mingyangchen, 	siliang\}@zju.edu.cn} 
% \and
% Imperial College London, London, United Kingdom \\
% \email{n.potyka@imperial.ac.uk}  \and
% Bosch Center for Artificial Intelligence, Renningen, Germany \\
% \email{trungkien.tran@de.bosch.com}  \and
% University of Southampton, Southampton, United Kingdom\\
 }

\maketitle              
\input{Sections/0_Abstract}
% typeset the header of the contribution

\input{Sections/1_Introduction}
\input{Sections/3_Preliminaries}
\input{Sections/4_Methodology}
\input{Sections/5_Experiment}
\input{Sections/2_Related_Work}
\input{Sections/6_Conclusion}

\section{Acknowledgments}
This work has been supported in part by the Zhejiang NSF (LR21F020004),  the NSFC (No. 62272411),  Alibaba-Zhejiang University Joint Research Institute of Frontier Technologies, and Ant Group.
\bibliographystyle{splncs04}
\bibliography{citation}
% \clearpage
% \input{Sections/7_Appendix}
\end{document}

%% file: Sections/0_Abstract.tex
\begin{abstract}
%Knowledge Graph Embedding (KGE) 
Knowledge graph embedding (KGE)
aims to map entities and relations of a knowledge graph (KG) into a low-dimensional and dense vector space via contrasting the positive and negative triples. In the training process of KGEs, negative sampling is essential to find high-quality negative triples since KGs only contain positive triples. Most existing negative sampling methods assume that non-existent triples with high scores are high-quality negative triples. However, negative triples sampled by these methods are likely to contain noise. Specifically, they ignore that non-existent triples with high scores might also be true facts due to the incompleteness of KGs, which are usually called false-negative triples. To alleviate the above issue, we propose an easily pluggable denoising mixup method called \textbf{\model}, which generates high-quality triples by refining sampled negative triples in a self-supervised manner. Given a sampled unlabeled triple, {\model} firstly classifies it into a marginal pseudo-negative triple or a negative triple based on the judgment of the KGE model itself. Secondly, it selects an appropriate mixup partner for the current triple to synthesize a partially positive or a harder negative triple. Experimental results on the knowledge graph completion task show that the proposed {\model} is superior to other negative sampling techniques, ensuring corresponding KGEs a faster convergence and better link prediction results.
\keywords{Knowledge graph embeddings \and Negative sampling \and Mixup.}
\end{abstract}

% 193词

% 2. 介绍KGE，使用正负样本对比来学习，并指出负样本质量对模型学到的表示质量影响很大
% 3. 说明目前负采样方法的缺陷：1.false negative,会对模型有很大的影响
% 4. 首先argue
% false ns 确实有很大影响
% 引出我们的方法adaptive mixup来合成样本，根据当前模型的预测结果讲伪负样本生成为部分标签正的样本，讲简单的负样本生成更难的负样本

%% file: Sections/1_Introduction.tex
\section{Introduction}
\label{introduction}
\begin{figure}[tb]
    \begin{minipage}{0.45\linewidth}
        \centering
        \centerline{\includegraphics[height=5.2cm]{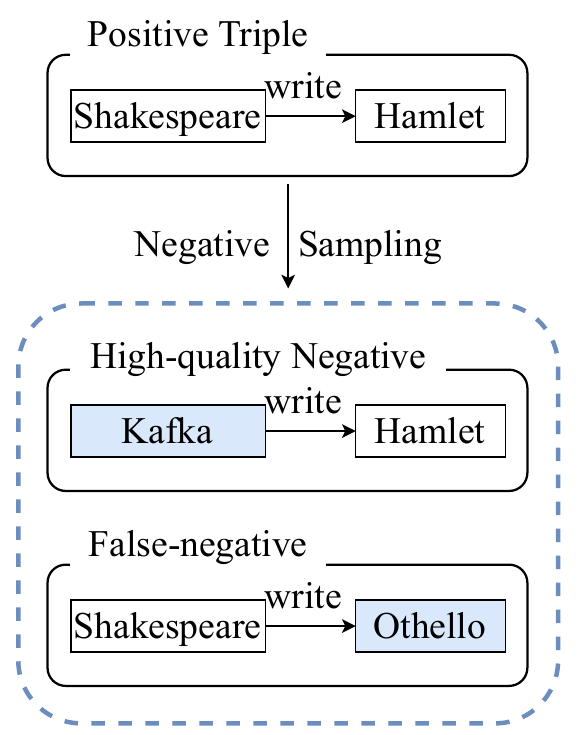}}\centerline{(a)}
    \end{minipage}%
    \begin{minipage}{0.45\linewidth}
        \centering
        \centerline{\includegraphics[height=5.2cm]{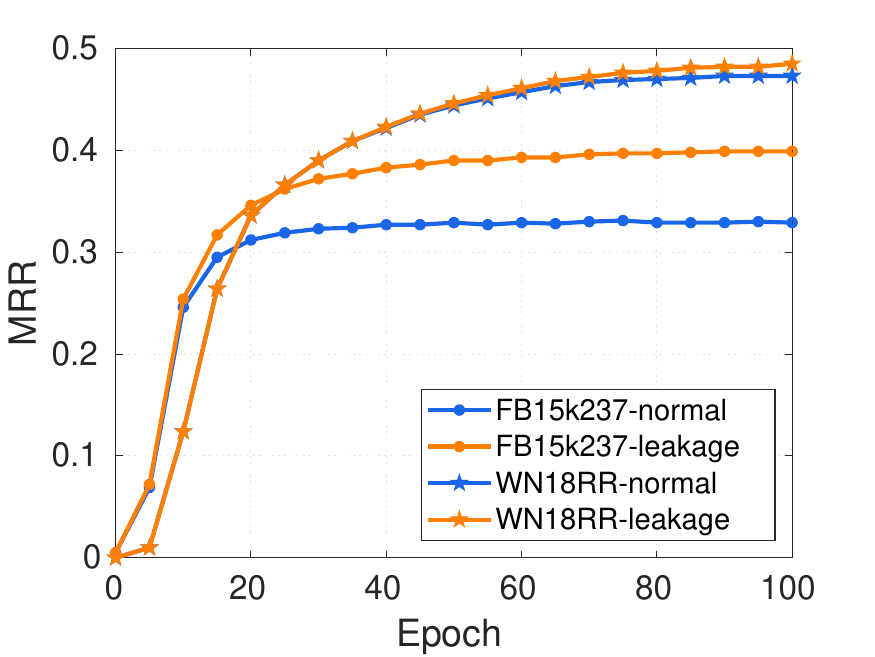}}\centerline{(b)}
    \end{minipage}%
	\caption{(a) The example of sampling false-negative triples. (b) Testing MRR performance v.s. Epoch based on RotatE. Normal means self-adversarial negative sampling. Leakage means ensuring sampled negative triples are not contained in the validation or test set. We regard triples in the validation and test sets as false-negative triples during training.} 
	\label{fig:example} 
% 	\vspace{-0.3cm}
\end{figure} 
Recently, knowledge graphs (KGs) have been successfully profitable in many practical applications, including question answering \cite{qa}, information retrieval \cite{xiong-2017-explicit} and dialogue systems \cite{graphdialog}. However, the KGs constructed manually or automatically still suffer from incompleteness.
Thus, completing KGs through efficient representation learning has been a hot topic. For flexible and efficient KG representation learning, knowledge graph embedding (KGE)\cite{TransE,XTransE} aims to represent entities and relations in KGs with 
real-valued vectors, also called embeddings.
% For example, TransE \cite{TransE} and its variants \cite{TransH,XTransE,TransR} use the relation for translating a head entity to a tail entity in vector space, successfully modeling the inversion and composition patterns among relations. 
KGEs have shown promising performance in KG related tasks, such as triple classification \cite{Ji_2022} and link prediction \cite{RotatE}.
They generally follow the same training paradigm. Specifically, they define 
a score function to measure the plausibility of triples through calculation with entity and relation embeddings, then learn the embeddings with the training objective of enlarging the gap between the scores of the positive and negative triples. 

Since KGs only contain positive triples, negative sampling methods for KGEs usually regard non-existent triples as negative triples. Formally, given a fact $(h,r,t)$ of a KG $\mathcal{G}$, negative sampling methods sample an entity $e$ among all candidate entities and replace the head entity $h$ or tail entity $t$ with $e$ to form a corrupted triple $(e, r, t) \notin \mathcal{G}$ or $(h,r,e) \notin \mathcal{G}$ as a negative triple. Previous works \cite{igan,zhang2019nscaching} have proved the quality of negative triples significantly affects the performance of KGEs, such as low-quality negative triples can cause gradient vanishing problems\cite{igan}.
Therefore, searching for high-quality negative triples is not only necessary but also vital for learning KGEs. 

Many negative sampling methods \cite{zhang2019nscaching,KBGAN} assume that non-existent triples with high scores are high-quality negative triples. These methods optimize the mechanism of negative sampling from different perspectives to search for non-existent triples with high scores. For example, KBGAN\cite{KBGAN} and IGAN\cite{igan} introduce a generative adversarial network (GAN)\cite{gan} to generate negative triples with high scores, and NSCaching\cite{zhang2019nscaching} introduces a caching mechanism to store corrupted triples with large scores, etc. 
However, these negative sampling methods for KGEs neglect the issue of sampling noisy triples, especially when they regard non-existing triples with high scores as high-quality negative triples. 
Because those corrupted triples with high scores might also be true facts with high probability due to the complementary capability of KGEs, which are usually called false-negative triples. As shown in Figure \ref{fig:example}(a), given a positive triple (\textit{Shakespeare}, \textit{write}, \textit{Hamlet}), the false-negative triple such as (\textit{Shakespeare}, \textit{write}, \textit{Othello}) may be sampled, which will give imprecise supervision signals to KGE models' training. Furthermore, we set up a toy experiment \footnote{We use the official open source code of RotatE model with the best config parameters: \url{https://github.com/DeepGraphLearning/KnowledgeGraphEmbedding}} to quantify the effect of false-negative triples on KGEs' training. As shown in Figure \ref{fig:example}(b), compared with normal negative sampling, negative sampling with data leakage can improve MRR by 21.3{\%}
and 1.2{\%} on FB15K237 and WN18RR respectively. This phenomenon indicates that although the probability of sampling false-negative triples is very low, the false-negative triples do mislead the learning of KGE and degrade the inference ability of KGE models.  

% due to the incompleteness of the KG, especially when they regard non-existing triples with high scores as high-quality negative triples. Because those triples might also be correct with high probability due to the complete capability of KGEs. As shown in Figure \ref{fig:example}(a), given a positive triple (\textit{Shakespeare}, \textit{write}, \textit{Hamlet}), the false-negative triple such as (\textit{Shakespeare}, \textit{write}, \textit{Othello}) may be sampled, which will give imprecise supervision signals to KGE models' training. Furthermore, we set up a toy experiment \footnote{We use the official open source code of RotatE model with the best config parameters: \url{https://github.com/DeepGraphLearning/KnowledgeGraphEmbedding}} to quantify the effect of false-negative triples on KGEs' training. As shown in Figure \ref{fig:example}(b), although the probability of sampling false-negative triples is very low, the false-negative triples do mislead the learning of KGE and degrade the inference ability of KGE models. Meanwhile, the degree of the effect is related to the denseness of KG itself. 
% 这里想说的是1.之前的方法忽视了false-negatives
% 2.并且进一步说明false-negatives影响很大，所以忽视它是个需要关注的问题，体现它的价值

Therefore, it is important to consider the challenging denoising problem when sampling high-quality negative triples. In this paper, to address the above issue, we propose a novel and easily pluggable framework \textbf{\model} which could generate high-quality triples that are beneficial for models' training by refining negative triples. Specifically, {\model} contains two modules, namely Marginal Pseudo-Negative triple Estimator (MPNE) and Adaptive Mixup (AdaMix). Given sampled corrupted triples $(h,r,e)$ or $(e, r, t)$, the MPNE module firstly leverages the current predictive results of the KGE model to estimate whether these corrupted triples contain noisy triples, then divides them into marginal pseudo-negative triples which more likely contain noisy triples and true-negative ones which are more likely negative triples. Then, in order to refine corrupted triples, the AdaMix module selects an appropriate mixup partner $e^\prime$ for $e$, then mixes them in the entity embedding space. Overall, {\model} generates partially positive triples for marginal pseudo-negative triples and harder negative triples for true negative ones as high-quality triples to help the training of KGEs.

% Specifically, given a sampled negative triple, $(h,r,e)$ or $(e, r, t)$, {\model} firstly leverages the current predictive power of the KGE model to estimate whether this triple is a marginal pseudo-negative triple, namely likely to be a false negative one or a true-negative one.
% Then, in order to effectively use label information of false-negative triples, {\model} selects an appropriate mixup partner $e^\prime$ for $e$, then mixes them in the entity embedding space.  Finally, {\model} 
% % can 
% generate\wen{s} partially positive triples for marginal pseudo-negative triples and high-quality negative triples for true negative ones. It does not mind sampling low-quality negative triples and false-negative triples since the mixup mechanism as a way of augmentation could help
% generate more informative triples with correct labels for KGE models' training.
In summary, the contributions of our work are as follows: 
% (这里想说明我们方法的优点是自监督+可插拔)
\begin{itemize}
    \item We propose a simple and efficient denoising framework, named \textbf{\model}. It generates high-quality triples by refining sampled negative triples without additional information.
    % \item We design two pluggable modules, which can be combined with other negative sampling methods. Meanwhile, it does not waste training time on additional parameters.
    \item We design two pluggable modules MPNE and AdaMix, which 
    are general to be combined with other negative sampling methods and
    are efficient for not wasting training time on additional parameters.
    \item We conduct extensive experiments on two benchmark datasets to illustrate the effectiveness and the superiority of our whole framework and each module.
\end{itemize}

%% file: Sections/3_Preliminaries.tex
\section{Preliminaries}
\subsubsection{Notations}
We use lower-case letters $h$, $r$, and $t$ to represent the head entity, relation, and tail entity in triples respectively, and
the corresponding boldface lower-case letters $\textbf{h}$, $\textbf{r}$ and $\textbf{t}$ indicate their embeddings. Let $\mathcal{E}$ and $\mathcal{R}$ represent the set of entities and relations respectively, we denote a knowledge graph as $\mathcal{G} = \{ \mathcal{E}, \mathcal{R}, \mathcal{T}\}$ where $\mathcal{T} = \{(h,r,t) \}\subseteq \mathcal{E} \times \mathcal{R} \times \mathcal{E}$ is the set of facts in KG. For training of KGEs, a dataset $\mathcal{D}$ of triples with labels is used $\mathcal{D} = \{ ((h_i,r_i,t_i), y_i)\}_{i=1}^{n_p + n_u} $, which includes $n_p$ positive triples with label $y=1$ and $n_u$ negative triples with label $y=0$. $n_p$ usually is equivalent to the size of fact set that $n_p = |\mathcal{T}|$. Since negative triples are not included in $\mathcal{G}$, thus they are usually created by negative sampling methods.
\subsubsection{Negative Sampling for KGE}
Unlike the negative sampling at the instance level in the computer vision domain~\cite{incremental}, the negative sampling for KGE is sampling and replacing entities in facts.
More specifically, based on the fact set $\mathcal{T}$, the negative sampling for KGE sample entities $e \in \mathcal{E}$ and replace either $h$ or $t$ in facts with $e$ to construct a set of corrupted triples $\mathcal{T}_u$ which is computed as follows:
\begin{equation}
\mathcal{T}_u = \bigcup_{(h,r,t)\in \mathcal{T}}\{(e, r, t) \notin \mathcal{T}\} \cup \{(h, r, e) \notin \mathcal{T}\}.
 \label{corrupt_triples}
\end{equation}
Following the closed-world assumption \cite{Ji_2022}, the majority of negative sampling methods for KGE treat all non-existent triples as negative triples. Thus,  
\begin{equation}
\begin{split}
    \mathcal{D} = &\{((h, r, t), y = 1)| (h, r, t) \in \mathcal{T}\} 
    \\
    &\cup\{((h',r,t'), y = 0)| (h', r, t') \in \mathcal{T}_u\}.
    \\
\end{split}    
\end{equation}
Therefore, the optimization objective for KGE models learning based on closed-world assumption(CWA) can be formulated as follows~\cite{RotatE}:

%  & -\sum_{((h,r,t)\in\mathcal{T}}
\begin{equation}
 \mathcal{L} =
 -\log\sigma(f(h,r,t))
 -\sum_{(h', r, t') \in \mathcal{T}_u}\log \sigma(-f(h', r, t')),
 \label{norm_loss}
\end{equation}
where $f(h,r,t)$ denotes the score function of KGEs to assess the credibility of $(h,r,t)$.
\subsubsection{KGE Score Function}

The two most typical score functions for a triple $(h,r,t)$ are:

(1) The \textbf{distance}-based score function which evaluates the score of triples based on the Euclidean distance between vectors, such as the score function of  RotatE~\cite{RotatE}:
\begin{equation}
    f(h, r, t) = \gamma - \Vert \textbf{h} \circ \textbf{r} - \textbf{t} \Vert,
    \label{eq4}
\end{equation}
where $\gamma$ is the margin, $\circ$ indicates the hardmard product.

(2) The \textbf{similarity}-based score function which evaluates the score of triples based on dot product similarity between vectors, such as the score function of  DistMult~\cite{DistMult}:
\begin{equation}
    f(h, r, t) = \textbf{h}^\top diag(\textbf{M}_r) \textbf{t},
    \label{eq5}
\end{equation}
where $diag(\textbf{M}_r)$ represents the diagonal matrix of the relation $r$.

%% file: Sections/4_Methodology.tex
\section{Methodology}
\label{sec:method}
\begin{figure*}[t]
    \centering
\includegraphics[width=1.0\linewidth]{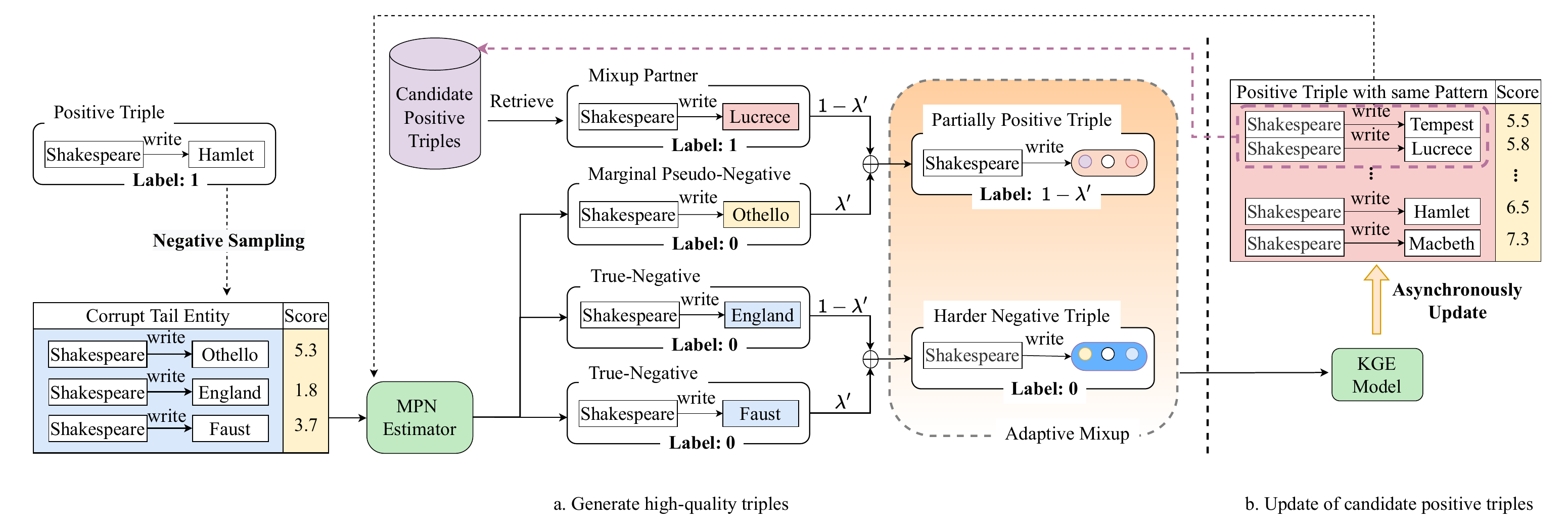}
	\caption{An overview of the {\model} framework based on corupting tail entity.}
	\label{fig:overview}
\end{figure*}

In this section, we introduce the proposed method \textbf{\model},
which is a novel and easily pluggable framework for generating high-quality triples. Recalling the denoising challenge in sampling high-quality negative triples, we design two modules to address the above challenge, namely Marginal Pseudo-Negative Triple
Estimator (MPNE) and Adaptive Mixup(AdaMix) modules. The MPNE module leverages the current predictive results of the KGE models to divide unlabeled corrupted triples into pseudo-negative triples and true-negative triples. Then, the AdaMix module selects a suitable mixup partner for each corrupted triple and mixes them in the entity embedding space to generate partially correct triples or harder negative triples to help train the KGE model. The overview of {\model}
is shown in Figure \ref{fig:overview}.

% As shown in \cref{fig:overview}, the whole pipeline consists of two components: marginal pseudo-negative estimation (MPNE) module and adaptive mixup (AdaMix) module. Given corrupted triples from uniform sampling, the MPNE module divides them into pseudo-negative triples and negative triples according to the current predictive results of the KGE model (\cref{method:mpne}), where pseudo-negative triples are more likely to be positive rather than negatives ones. Then, the AdaMix module selects suitable mixup partners for the corrupted triples according to the classification results from MPNE
% and mixes them in the entity embedding space to generate partially correct triples or high-quality negative triples to help train the KGE model (\cref{method:adamix}).
% Through MPNE and AdaMix, {\model} could generate high-quality triples given corrupted triples of any quality, thus does not have the drawbacks of inefficient searching and neglecting false-negative triples as previous negative sampling methods.

% #---------------------------------
% #---------------------------------

\subsection{Marginal Pseudo-Negative Triple
Estimator (MPNE)}

\label{method-mpne}
Motivated by sampled corrupted triples with high scores more likely to contain noise due to the incompleteness of KGs, we aim to leverage the predictive results of the KGE model itself to recognize noisy triples. In this work, we divide sampled corrupted triples into two subsets, such as marginal pseudo-negative triples which are likely to contain noise according to the current predictive results of the KGE model, and true-negative triples which are more likely negative triples. Because the KGE model has different learning levels for each relation pattern due to the long-tail distribution in KGs\cite{long-tail} and the phenomenon that the discriminative ability of the KGE model is continuously improving during the training process. We need to design a module to dynamically estimate noisy triples. Considering negative sampling for KGE is to replace the head entity or tail entity in a positive triple, the unreplaced binary terms in a positive triple are invariant. We call the invariant binary terms a pattern. So $\mathcal{T}_{u}$ can be reformulated as follows:
\begin{equation}
 \label{corrupt_pair}
\begin{split}
 \mathcal{T}_u = &\{(e, pattern)|e \in \mathcal{E}, pattern=(r,t)\} \\
 &\cup \{(pattern, e)|e \in \mathcal{E}, pattern=(h,r)\}.
\end{split}
\end{equation}
The set of positive triples with the same pattern ((\textit{h}, \textit{r}) or (\textit{r}, \textit{t})) as follows:
\begin{equation}
 \label{t_pair}
\begin{split}
 &\mathcal{T} = \mathcal{T}_{pattern}^{rt} \cup \mathcal{T}_{pattern}^{hr},
 \\
 &\mathcal{T}_{pattern}^{rt} = \{(e_i, r,t) \in \mathcal{T}|e_i \in \mathcal{E}\}, 
 \\
 &\mathcal{T}_{pattern}^{hr} = \{(h,r,e_j) \in \mathcal{T}|e_j \in \mathcal{E}\}.
\end{split}
\end{equation}
Then we treat corrupted triples whose scores are close to the scores of the positive samples with the same pattern as marginal pseudo-negative triples $\mathcal{T}_{mpn}$.
% By collecting the scores of positive triples with the same pattern, we can estimate the current learning level of the KGE model for this pattern. Notably, the KGE model has different learning levels for each pattern, due to the long-tail distribution in KGs. Then 
% we treat negative triples that confuse the current KGE model as noisy triples. In other words, the KGE model assumes that these triples are likely to be positive triples. We refer to the noisy triples estimated in this way as marginal pseudo-negative triples $\mathcal{T}_{mpn}$, which are likely to be positive triples. 
So $\mathcal{T}_{mpn}$ can be formulated as follows:
\begin{equation}
\label{mpn}
\begin{split}
 &\mathcal{T}_{mpn}=  \{((h^{\prime}, r, t^{\prime}), y=0)|  (h^{\prime}, r, t^{\prime}) \in \mathcal{T}_{{u}}, \\
 & -\delta_{T} + 
 [f^{\mathcal{T}_{pattern}}]_{min}
 \leqslant f(h^{\prime}, r, t^{\prime})\leqslant [f^{\mathcal{T}_{pattern}}]_{mean}\},
\end{split}
\end{equation}
where $f^{\mathcal{T}_{pattern}}$ is the collection of scores of positive triples with the same pattern, i.e.,  $f^{\mathcal{T}_{pattern}} = \{f(h,r,t)| (h,r,t) \in \mathcal{T}_{pattern} \}$, and  $[X]_{min}$ and $[X]_{mean}$ is the minimum and mean value of $X$. $\delta_{T}$ is a hyper-parameter controlling the estimation range at the T-th training epoch. Specifically, $\delta_{T} = \delta \cdot min(\beta, T/T_{0})$, where $T_{0}$ denotes the threshold of stopping increase, $\delta$ and $\beta$ are hyper-parameters.
Notably, inspired by the characteristics of complex relations in KG, namely 1-N, N-1, 1-1, and N-N defined in TransH~\cite{TransH}, we record the num of positive triples with the same pattern.
When $|\mathcal{T}_{pattern}|$ is lower, the probability that the corrupted triples based on this pattern are noisy triples is lower. So after sampling negative triples $\mathcal{T}_{u}$, we 
set a threshold $\mu$ to decide whether to estimate the set $\mathcal{T}_{mpn}$ from $\mathcal{T}_{u}$ as follows:
\begin{equation}
 \mathcal{T}_{{u}}=
\begin{cases}
\mathcal{T}_{mpn} \cup \mathcal{T}_{\widetilde{u}}& if        {|\mathcal{T}_{pattern}| >= \mu},\\
\mathcal{T}_{\widetilde{u}}& if  {|\mathcal{T}_{pattern}| < \mu},
\end{cases}
\label{pair_num}
\end{equation}
where $\mathcal{T}_{\widetilde{u}}$ contains triples which are regarding as true-negative triples.
\subsection{Adaptive Mixup(AdaMix)}
\label{method:adamix}
To address the denoising challenge in sampling high-quality negative triples, an intuitive approach is to directly label the triples in $\mathcal{T}_{mpn}$ as 1. However, when the KGE model does not have the strong distinguishable ability, especially in the early training stage.
this approach can give the wrong supervisory signal to the KGE model. Inspired by the recent progress of mixup~\cite{MixGCF2021}, we aim to adapt the mixup technique to alleviate the wrong supervisory signal issue. Since the triples in $\mathcal{T}_{mpn}$ are more likely to be positive triples than $\mathcal{T}_{\widetilde{u}}$, we develop an adaptive mixup mechanism to guide the selection of a suitable mixup partner for each corrupted triple, to generate high-quality triples containing rich information. In specific, we first build a pool of candidate positive triples $\mathcal{T}_{cap}$ for each pattern from $\mathcal{T}_{pattern}$. $\mathcal{T}_{cap}$ contains the positive triples around the current learned boundary of the KGE model. $\mathcal{T}_{cap}$ can be formulated as follows:
\begin{equation}
\label{cap}
\begin{split}
 \mathcal{T}_{{cap}} = \{((h, r, t), y=1)|(h, r, t)  \in \mathcal{T}_{pattern}, 
 \\
 f(h,r,t)  \leqslant [f^{\mathcal{T}_{pattern}}]_{mean}\}.
\end{split}
\end{equation}
Since the embeddings of the KGE model are continually updated, we thus propose to update $\mathcal{T}_{{cap}}$ every epoch. 
Then, we select a mixup partner with the same pattern for each corrupted triple in $\mathcal{T}_{u}$. For marginal pseudo-negative triples in $\mathcal{T}_{mpn}$, which are more likely to be positive but annotated by negative, we uniformly choose a mixup partner from $\mathcal{T}_{cap}$ to generate a partially positive triple to take more precise supervision to the KGE model. Besides, for true-negative triples in $\mathcal{T}_{\widetilde{u}}$, we select another true-negative triple as a mixup partner to construct harder negative triples using a non-existent mixing entity in KG. It is worth noting that since the pattern of the mixup partner is the same as the pattern of the corrupted triple, the actual mixing object is mixing between corrupted candidate entity and the entity in the corresponding position of the mixup partner. Let $\mathbf{e}_{i}$, $\mathbf{e}_{j}$ denote the entity to be mixed in the corrupted triple and the corresponding mixup partner respectively, the overall mixup partner selection is formulated as follows:
\begin{equation}
\left(\mathbf{e}_{j}, y_{j}\right) \sim \begin{cases}\text { Uniform }\left(\mathcal{T}_{cap}\right) & \text { if }\left(\mathbf{e}_{i}, y_{i}\right) \in \mathcal{T}_{mpn}, \\ \text { Uniform }\left(\mathcal{T}_{\widetilde{u}} \right) & \text { if }\left(\mathbf{e}_{i}, y_{i}\right) \in \mathcal{T}_{\widetilde{u}}.
\end{cases}
\label{select_partner}
\end{equation}
Finally we mix each corrupted triple $(h^{\prime}_{i}, r_{i}, t^{\prime}_{i}) \in \mathcal{T}_{mpn} \cup \mathcal{T}_{\widetilde{u}}$ with its corresponding mixup partner $(h_{j},r_{j}, t_{j})$ in the entity embedding space of the KGE model to generate an augmented triple set $\widehat{\mathcal{T}}_{u}$.
Motivated by the modified mixup operator \cite{mixmatch}, the mixup operation is as follows:
\begin{equation}
\label{mixup1}
\begin{split}
\widehat{\mathbf{e}}_{i}&=\lambda^{\prime} \mathbf{e}_{i}+\left(1-\lambda^{\prime}\right) \mathbf{e}_{j}, \quad \widehat{y}_{i}=\lambda^{\prime} y_{i}+\left(1-\lambda^{\prime}\right) y_{j},\\
\lambda^{\prime}&=\max (\lambda, 1-\lambda),
\lambda \sim \operatorname{Beta}(\alpha, \alpha), \alpha \in(0, \infty),\\
\end{split}
\end{equation}
where $\lambda^{\prime}$ is a balance parameter. $\lambda^{\prime}$ can guarantee that the feature of each augmented entity $\widehat{\mathbf{e}}_{i}$ is closer to $\mathbf{e}_{i}$ than the mixup partner $\mathbf{e}_{j}$.

\begin{algorithm}[ht]
	\caption{Training procedure of {\model}}
	\label{alg:algorithm}
	\textbf{Input}: training set $\mathcal{P}=\{(h, r, t)\}$, entity set $\mathcal{E}$, relation set $\mathcal{R}$, embedding dimension $d$, scoring function $f$, the size of epoches E, the size of warm-up epoches W.\\
	%\textbf{Parameter}: Optional list of parameters\\
	\textbf{Output}: embeddings for each $e \in \mathcal{E}$ and $r \in \mathcal{R}$
	\begin{algorithmic}[1] %[1] enables line numbers
		\STATE \textbf{Initialize} embeddings for each $e \in \mathcal{E}$ and $r \in \mathcal{R}$; \\
		\STATE Get $\mathcal{T}_{pattern}$ from $\mathcal{T}$;
		\WHILE{epoch $<$ W}
		\STATE Warm up model using uniform sampling with K negative samples;
		\ENDWHILE
		\WHILE{epoch $<$ E}
		\STATE Sample a mini-batch  $\mathcal{T}_{\text {batch }} \in \mathcal{T}$;
		\WHILE{$(h, r, t) \in \mathcal{T}_{\text {batch }}$}
		\STATE Uniformly sample $M$ entities from $\mathcal{E}$ to form unlabeled corrupted triplets $\mathcal{T}_{u}=\{(h_{m}^{'}, r, t_{m}^{'}), m=1,2...M\}$; \\
		\STATE Estimate marginal pseudo-negative triples $\mathcal{T}_{mpn}$ using Eq.(\ref{pair_num})(\ref{mpn});
		\STATE Select the mixup partners for each corrupted triples using Eq.(\ref{select_partner});
		\STATE Construct $\widehat{\mathcal{T}}_{u}$ applying Eq.(\ref{mixup1}) to corrupted triples and their mixup partners;
		\STATE calculate loss functions using Eq.(\ref{mixup_loss}), then update the embeddings of entites and relations via gradient descent;
		\ENDWHILE
		\STATE Update $\mathcal{T}_{cap}$ using Eq.(\ref{cap});
		\ENDWHILE
	\end{algorithmic}
\end{algorithm}
\subsection{Traning the KGE Model}
\label{training}
Because our framework is pluggable, we can combine other negative sampling methods to train the KGE model. Here we show the training objectives based on the uniform negative sampling~\cite{TransE} and self-adversarial sampling~\cite{RotatE}. The loss function based on uniform sampling is as follows:
% \sum_{(h,r,t) \in {\mathcal{T}}}
\begin{equation}
\begin{aligned}
\mathcal{L}&= \ell(f(h,r,t), 1) + \sum_{(\widehat{h}_i,r,\widehat{t}_i) \in \widehat{\mathcal{T}}_{u}} \ell(f(\widehat{h}_i,r,\widehat{t}_i), \widehat{y}_{i}).
\end{aligned}
\label{mixup_loss}
\end{equation}
The loss function based on self-adversarial sampling is as follows:
% \mathcal{L}&= \sum_{(h,r,t) \in {\mathcal{T}}
\begin{equation}
\begin{aligned}
\mathcal{L}
& = \ell(f(h,r,t), 1)
\\& + \sum_{(\widehat{h}_i,r,\widehat{t}_i) \in \widehat{\mathcal{T}}_{u}}p(\widehat{h}_i,r,\widehat{t}_i) \ell(f(\widehat{h}_i,r,\widehat{t}_i), \widehat{y}_{i}),
\end{aligned}
\end{equation}
where  $\widehat{\mathcal{T}}_{u}
 =\operatorname{AdaptiveMixup}\left(\mathcal{T}_{u}, \mathcal{T}_{pattern}, \alpha\right)$ and $\ell(.,.)$ is the cross-entropy loss. Movever, the probability distribution of sampling high-quality triples is as follows: 
\begin{equation}
    p(\widehat{h}_j,r,\widehat{t}_j|\{(\widehat{h}_i,r,\widehat{t}_i)\})=\frac{\exp \alpha_{t} f(\widehat{h}_j,r,\widehat{t}_j))}{\sum_i \exp \alpha_{t} f(\widehat{h}_i,r,\widehat{t}_i)},
\end{equation}
where $\alpha_{t}$ is the temperature of sampling. The full training procedure is shown in Algorithm~\ref{alg:algorithm}

%% file: Sections/5_Experiment.tex
\section{Experiment}
\label{sec:experiment}
In this section, we perform detailed experiments to demonstrate the effectiveness of our proposed framework {\model} by answering the following questions. \textbf{Q1}: Does {\model} can mitigate the noisy triples issue? \textbf{Q2}: Whether {\model} can be effectively plugged into other negative sampling methods? \textbf{Q3}: How does each of designed modules influence
the performance of {\model}?

\input{Tables/Statistics}
% We firstly present the dataset statistics, baseline models for comparison, evaluation protocol, and implementation details. Then,
% we demonstrate the superiority of our method compared with several baselines on the link prediction task. Furthermore, we conduct further experiments, including the ablation study and the further anaylsis, to validate the effectiveness of each module of \model.
\subsection{Experiment Settings}

\subsubsection{Datasets} Two public datasets are utilized for experiments, including WN18RR \cite{dettmers2018conve} and FB15K237 \cite{fb15k237}. WN18RR is a subset of WN18 \cite{TransE}, where inverse relations are
deleted. Similarly, FB15k237 is a subset of FB15K \cite{TransE}, which comes from FreeBase \cite{freebase}. FB15K237 is denser than WN18RR, so it is more affected by false-negative triples. The statistics of the two datasets are given in Table \ref{statistic}.

\subsubsection{Baseline methods}
% To verify the effectiveness of our NSGenerating method, 
We compare {\model} to seven negative sampling baselines. The details of baseline methods are presented as follows:
\begin{itemize}
    \item \textit{Uniform Sampling} \cite{TransE}.The basic negative sampling method, which samples negative triples from a uniform distribution.
    \item \textit{Bernoulli Sampling} \cite{TransH}.
which sample negative triples from a Bernoulli distribution considering false-negative triples.
%     \item \textit{KBGAN} \cite{KBGAN}
% The KBGAN tries to use generative adversarial networks in the negative sampling process.
    \item \textit{NSCaching} \cite{zhang2019nscaching}.
The NSCaching introduces the cache strategy as a general negative sampling scheme.
    \item \textit{Self-adversarial Sampling} \cite{RotatE}.
It utilizes a self-scoring function and samples negative triples according to the current embedding model.
    \item \textit{RW-SANS} \cite{sans}.
It samples negative triples from the k-hop of the node neighborhood by utilizing the graph structure.
    \item \textit{CANS} \cite{niu2022cake}.
The CANS is a component of CAKE~\cite{niu2022cake} responsible for solving the invalid negative sampling challenge. Considering our method focuses on negative sampling for KGEs, we mainly compare CANS instead of CAKE.
    \item  \textit{ESNS} \cite{esns}.
    It takes semantic similarities among entities into consideration to tackle the issue of false-negative samples. 
\end{itemize}
\subsubsection{Evaluation protocol}
Following the previous work \cite{niu2022cake}, we calculate the score of triples in the test dataset by employing the learned KG embeddings and the score function. Then we can get the rank of the correct entity for each test triple based on the filtered setting, where all corrupted triples in the dataset are removed. The performance is evaluated by four metrics: mean reciprocal rank (MRR) and Hits at 1,3,10 (Hits@N). Higher MRR and Hits@N mean better performance.

\input{Tables/experiment}

\input{Tables/experiment2}
\subsubsection{Implementation details}
We firstly use the negative sampling method to warm-up\footnote{We will discuss warm-up in the ablation study} 8 epochs to give the KGE model discriminative ability, and then use our method to generate high-quality triples for further training. We use PyTorch \cite{pytorch} framework to implement our method and Adam \cite{adam} optimizer for model training. The mixup balance hyperparameter $\alpha$ is fixed to 1. In addition, we use grid search to tune hyper-parameters on the validation dataset. Specifically, we choose $\beta$ from $\{1, 3\}$, $\delta$ from $\{0, 0.1, 1\}$, $\mu$ from $\{1,3,5\}$. The learning rate is tuned from 0.00001 to 0.01. The margin is chosen from $\{3, 6, 9, 12\}$. To make a fair comparison, the negative sampling size of each baseline is the same as generated negative triples size in our method. Specifically, for translational distance-based models, we follow the experimental setup in CANS~\cite{niu2022cake}. The negative sampling size is 16. For semantic matching-based approaches, we set the negative sampling size as 50 following ESNS\cite{esns}.

% During the whole training process, we follow the experimental setup in CANS~\cite{niu2022cake} for a fair comparison. Specifically, the embedding size and the batch size are the same as those of each basic model. The generated triples size in our method is set as 16, which is the same as negative sampling size of other methods.
% % The size of uniform sampling $K$ is fixed as 64. The warm up epoch is adjusted in $\{5, 8\}$.

\subsection{Experimental Results (Q1)}
We compare the overall performance of {\model} applied to different KGE models in the link prediction task to answer Q1. $X^*$, $X^\Box$, and $X^\Delta$ indicate the results are taken from \cite{zhang2019nscaching}, ~\cite{niu2022cake} and official code reproduction based on same setting respectively. First, we conduct experiments with translational distance-based KGE models. The link prediction results of TransE with different negative sampling methods on the two datasets are shown in Table \ref{results}. We can observe that {\model} can effectively improve the performance of the TransE model on each dataset. Compared with the best method without considering false-negative triples, our {\model} method improves MRR by 0.9\%, 7.8\% on WN18RR, and FB15K237. Even compared to CANS which reduces false-negative triples with commonsense, our method improves MRR by 6.7\% on FB15K237. Besides, the improvement of our method on WN18RR is not as pronounced as on FB15K237, which is consistent with the observation that false-negative triples have a lower effect on WN18RR than the degrading effect on FB15K237 in Figure \ref{fig:example}(b). Furthermore, we conduct experiments with recently proposed KGE models such as RotatE and HAKE on two datasets. From the results shown in Table \ref{reulst2}, our {\model} outperforms all the other negative sampling methods on FB15K237 dataset and achieves competitive results on WN18RR. Second, we conduct experiments with semantic matching KGE models. As shown in Table \ref{result3}, we can observe that our method DeMix outperforms ESNS, which specifically aim to tackle the issue of false-negative triples, on both datasets incorporating DistMult and ComplEx as backbones. These results demonstrate the superiority and effectiveness of our method. The results combined with different KGE models also illustrate the generalizability of our approach. 
\input{Tables/experiment3}

\input{Tables/combine}
\input{Tables/ablation}
\subsection{Combine with other NS (Q2)}
% \label{ablation}
To verify whether our approach is plug-and-play, we implement {\model} upon other negative sampling methods to answer Q2. From the results shown in Table \ref{tab:combine}, our proposed method can be effectively combined with different negative sampling methods, and all of them can obtain significant improvements. Such results demonstrate that {\model} can be a plug-and-play component for existing negative sampling methods.

\subsection{Ablation Study (Q3)}
\label{ablation}
We conduct ablation studies to show the contribution of different modules in {\model} to answer Q3.
We choose HAKE \cite{hake} model as the backbone since it has high performance. Specifically, we integrate our framework into HAKE based on the following three ablation settings: 1) removing warm up the KGE model (-WARM); 2) neglecting marginal pseudo-negative triples (-MPNE); 3) directly using the label information of noisy triples (-AdaMix). The results of ablation studies using HAKE on FB15K237 are shown in Table \ref{tab:ablation}. The results show that all ablation settings lead to degraded performance. In specific, we observe that the warm-up has a slight effect on the training of the KGE model. AdaMix module is essential for model performance, indicating the adaptive mixup mechanism can take more precise supervision to the KGE model. 
Moreover, we also find that the performance drops significantly after removing the MPNE module, which indicates the validity of estimating marginal pseudo-negative triples.

\subsection{Further Analysis}
\begin{figure}[t]
% \begin{minipage}{0.48\linewidth}\centering
% \centerline{\includegraphics[height=4.0cm]{Figures/conv.eps}} 
\begin{minipage}{\linewidth}\centering
\centerline{\includegraphics[height=5.5cm]{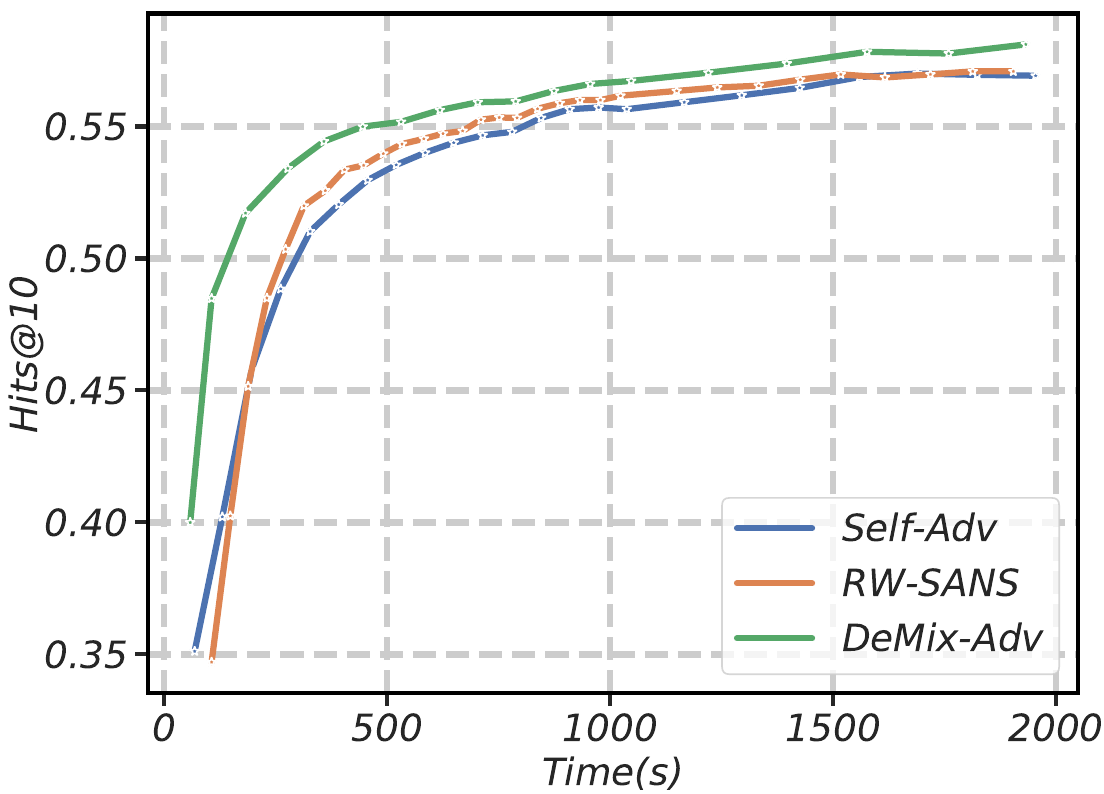}}
\end{minipage}
\centering
\caption{Evaluating Hits@10 performance v.s.  clock time (in seconds)  of HAKE based on WN18RR.}
\label{analysis1}
\end{figure}

\begin{figure}[t]
\begin{minipage}{\linewidth}\centering
\centerline{\includegraphics[height=5.5cm]{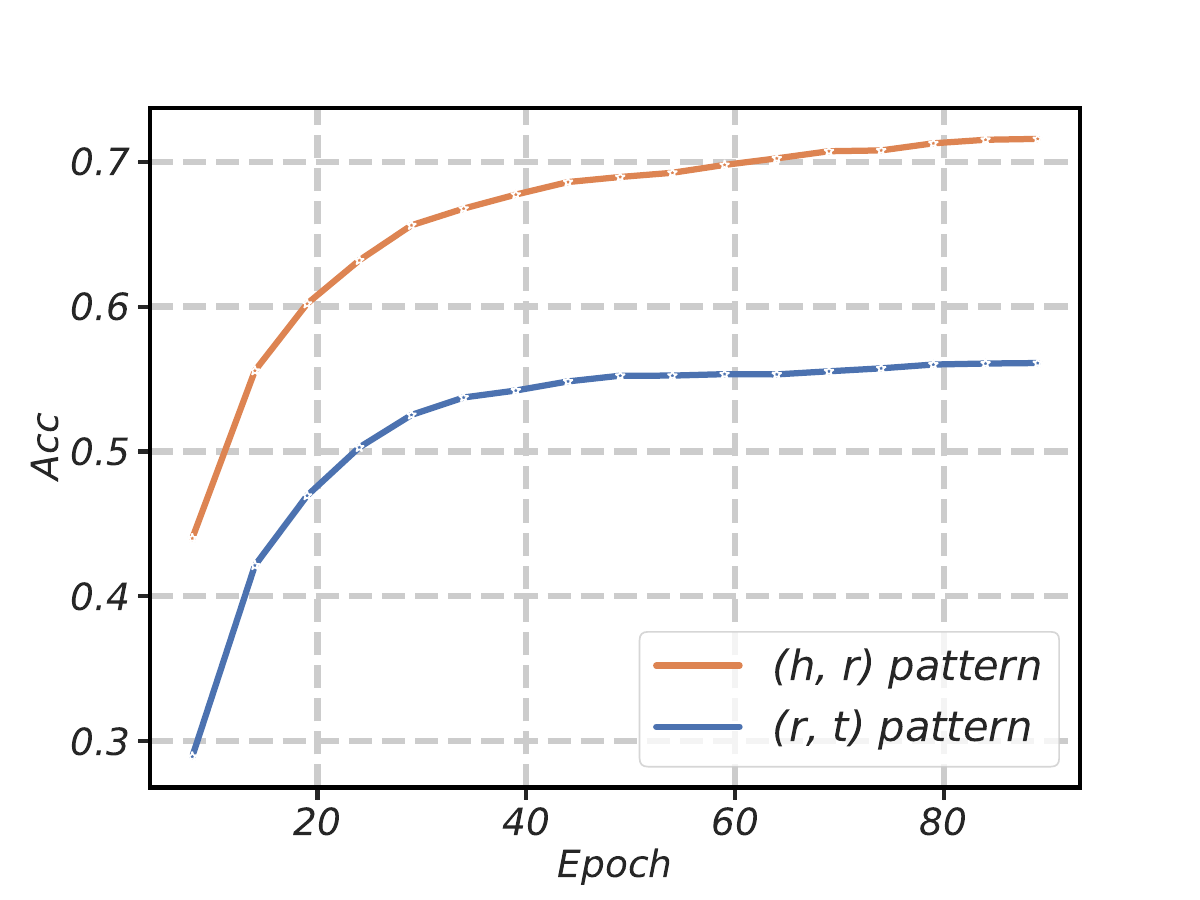}}
\end{minipage}
\centering
\caption{The estimating accuracy of the MPNE module of HAKE based on FB15K237 with warming up 8 epochs.}
\label{analysis2}
\end{figure}
% In this part, we do further analysis to illustrate whether our method addresses both inefficient searching and neglecting false-negative triples challenges.

\subsubsection{Convergence Speed}
First, we demonstrate the generating manner of {\model} can help the KGE model converge quickly. To ensure a clear observation,
we compare the convergence speed of our method with other negative sampling
methods using the same batch size and learning rate on WN18RR.
Figure \ref{analysis1} shows the convergence of evaluating Hits@10 of HAKE based on WN18RR. We can observe our method help the KGE model converge more quickly compared with other searching-based methods. This means that even on WN18RR where the effect of false-negative triples is small, our method can help the model converge faster by generating high-quality triples, while not wasting training time on additional parameters.

\subsubsection{Estimation Accuracy}
In order to investigate whether the MPNE module in {\model} can indeed identify false-negative triples, we inject triples from the validation set and test dataset into the MPNE module, then observe the estimation accuracy of the MPNE module for these false-negative triples as the training time increases. In specific, we calculate the estimation accuracy separately according to two patterns, i.e. (\textit{h}, \textit{r}) and (\textit{r}, \textit{t}). Especially if a pattern does not exist in $\mathcal{T}_{pattern}$, we do not estimate this triple based on this pattern. Finally, the number of false-negative triples are $28926$ and $34410$ based on $(h,r)$ pattern and $(r, t)$ pattern respectively. As shown in Figure \ref{analysis2}, After warming up HAKE 8 epochs, the estimation accuracy increases along the training, which implies our method can leverage the judgment power of the model itself to efficiently estimate false-negative triples. 

\subsubsection{Visualization of Entity Embeddings}
\begin{figure}[tb]
    \begin{minipage}{0.5\linewidth}
        \centering
        \centerline{\includegraphics[height=5.5cm]{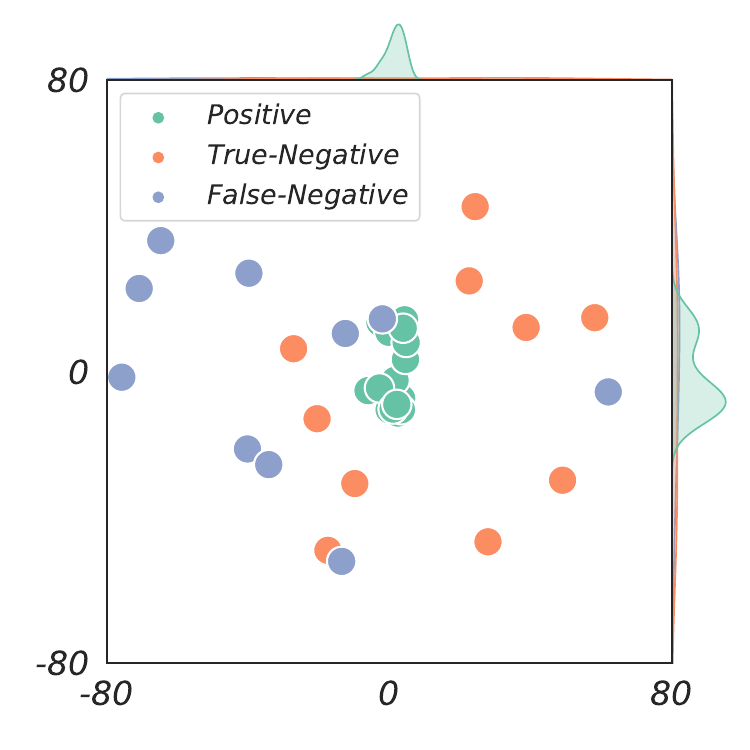}}\centerline{w/o {\model}}
    \end{minipage}%
    \begin{minipage}{0.5\linewidth}
        \centering
        \centerline{\includegraphics[height=5.5cm]{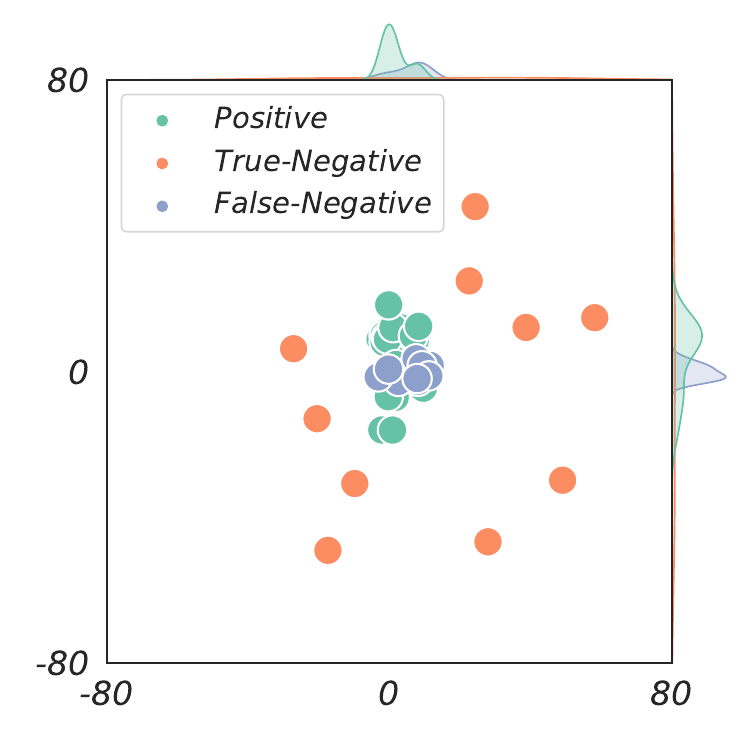}}\centerline{{\model}}
    \end{minipage}%
	\caption{2D t-SNE visualisation of tail entities with their embeddings of positive triples near the decision boundary, false-negative triples, and true-negative triples with the same pattern (\textit{marriage}, \textit{location}).} 
	\label{visualize} 
\end{figure}

We visualize entities embeddings to verify the validity of our adaptive denoising mixup mechanism. In specific, we random sample an input (\textit{h}, \textit{r}) pattern from FB15K237, namely (\textit{marriage}, \textit{location}), then retrieve 20 positive triples, in the training set, near the decision boundary, 10 false-negative triples, in the validation set and test set, with the lowest scores, and 10 true-negative triples with the highest scores. We visualize tail entities embeddings of these triples. From Figure \ref{visualize}, we can notice that our method can push the false-negative triples closer to the positive triples and away from the true-negative triples, which indicates our adaptive denoising mixup mechanism can alleviate the noisy triples issue.

%% file: Tables/Statistics.tex
\begin{table}
\setlength{\tabcolsep}{4.9mm}
 \centering
 \begin{tabular}{c|cc|ccc}
 \toprule
Dataset		& \#Rel	& \#Ent	 & \#Train	& \#Valid	& \#Test \\
 \midrule

 WN18RR   & 11           & 40,943            & 86,835   & 3,034    & 3,134 \\
 FB15K237   & 237           & 14,541            & 272,115   & 17,535    & 20,466 \\

 \bottomrule
 \end{tabular}
 \caption{Statistics of two benchmarks. \#Rel, \#Ent, represent the number of relations, and entities of each dataset, respectively.}
 \label{statistic}
 \end{table}

%% file: Tables/experiment.tex
\begin{table*}[t]
    
%\setlength{\abovecaptionskip}{-0.005cm}
%% increase table row spacing, adjust to taste
%\renewcommand{\arraystretch}{0.6}

\centering
\setlength{\tabcolsep}{2.6mm}
{
\footnotesize
\begin{tabular}{l|cc|cc}
\toprule
\multirow{2}*{Translational Distance-based Models} & \multicolumn{2}{c|}{WN18RR} & \multicolumn{2}{c}{FB15K237} \\
	& MRR	& Hits@10	& MRR	& Hits@10	\\
\hline
\multicolumn{5}{c}{Neglecting False-negative Triples}\\
\hline
TransE+Uniform          &  0.175$^*$    &  0.445$^*$ & 0.171$^\Box$    & 0.323$^\Box$    \\
% TransE+KBGAN            &0.211  &0.479  &0.381  &0.022   &0.277  &0.452  &0.312  &0.189          \\
% TransE+NSCaching        &      &      &      &                  &0.205 & 0.353    &0.226      & 0.131  \\
% TransE+KBGAN            &0.181  &0.432  &\textbf{0.293}  &\textbf{0.466 }         \\
TransE+NSCaching       &0.200$^*$     &0.478$^*$                       &0.205$^\Box$ & 0.353$^\Box$     \\
TransE+Self-Adv &0.215$^\Delta$ &\underline{0.516}$^\Delta$ & 0.268$^\Box$    & 0.454 $^\Box$   \\
% --------addtional information
TransE+RW-SANS$^\Delta$ &\underline{0.218}  &0.510  &0.295  &0.483            \\
% \midrule

% % \multicolumn{9}{c}{Considering false-negative triples}
% \midrule

\hline
\multicolumn{5}{c}{Considering False-negative Triples}\\
\hline
TransE+Bernoulli*  &0.178  &0.451  &0.256  &0.419       \\
TransE+CANS$^\Box$           & -     &  -        & \underline{0.298}   & \underline{0.490}   \\

% ---------------ours
\hline
\multicolumn{5}{c}{Ours}\\
\hline
% TransE+\textbf{NSG\_uni}	   & \textbf{0.209}   & \underline{0.499}         & \textbf{0.315}  & \underline{0.502}        \\
TransE+\textbf{{\model}-Adv}	   &\textbf{0.220}  &\textbf{0.521}   & \textbf{0.318}  & \textbf{0.510}       \\
\bottomrule
\end{tabular}}
\caption{Link prediction results for TransE on two datasets. \textbf{{\model}-Adv} denotes {\model} based on self-adversarial sampling, \textbf{Bold} numbers are the best results for each type of model. Underlined numbers mean the best performances of baselines.}
\label{results}
% \vspace{-0.3cm}
\end{table*}

%% file: Tables/experiment2.tex
\begin{table*}[htb!]
%\setlength{\abovecaptionskip}{-0.005cm}
%% increase table row spacing, adjust to taste
%\renewcommand{\arraystretch}{0.6}
\centering
\resizebox{\textwidth}{!}{
\begin{tabular}{l|cccc|cccc}

\toprule
\multirow{2}*{Translational Distance-based} & \multicolumn{4}{c|}{WN18RR} & \multicolumn{4}{c}{FB15K237} \\
	& MRR	& Hits@10	& Hits@3  & Hits@1	& MRR	& Hits@10	& Hits@3  & Hits@1\\
\midrule
RotatE+Uniform$^\Delta$ &0.471 &0.560 &0.488 &0.424 &0.282 &0.462 &0.314 &0.191                \\
RotatE+Self-Adv     &0.476$^\Delta$&0.570$^\Delta$&0.490$^\Delta$&0.428$^\Delta$                 & 0.269$^\Box$     & 0.452$^\Box$     & 0.298$^\Box$     & 0.179$^\Box$ \\

% -------additional information
RotatE+RW-SANS$^\Delta$ &\underline{0.478}  &\underline{0.572} &\underline{\textbf{0.494}}  &\underline{\textbf{0.430}} &0.295  &0.481  &0.327  &0.202                \\
RotatE+CANS$^\Box$	           & -     &  -    &  -    &    -          & \underline{0.296}    & \underline{0.486}    & \underline{0.329}	& \underline{0.202} \\

% #-----------ours 

% RotatE+\textbf{NSG\_uni}	  & \textbf{0.472}    & \textbf{0.563}    & \textbf{0.487}	& \underline{0.427}     & \underline{0.322}  & \underline{0.515}      & \underline{0.359}   & \underline{0.227}  \\
RotatE+\textbf{{\model}-Adv}	  & \textbf{0.479}    & \textbf{0.576}    &0.492	& 0.428     & \textbf{0.329}  & \textbf{0.518}      & \textbf{0.366}   & \textbf{0.235}  \\

\midrule
HAKE+Uniform$^\Delta$ &0.493 &0.580 &0.510 &0.450 &0.304 &0.482 &0.333 &0.216                \\
HAKE+Self-Adv      &\underline{0.495}$^\Delta$&\underline{0.580}$^\Delta$&\underline{0.513}$^\Delta$&\underline{0.450}$^\Delta$                & 0.306$^\Box$    & 0.486$^\Box$    & 0.337$^\Box$     & 0.216$^\Box$ \\

% -------additional information
HAKE+RW-SANS$^\Delta$  &0.492  &0.579  &0.507  &0.446   &0.305  &0.488   &0.336   &0.214                \\
HAKE+CANS$^\Box$	          & -     &  -    &  -    &    -           & \underline{0.315}  & \underline{0.501}      & \underline{0.344}   & \underline{0.221}  \\

% #-----------ours 
% HAKE+\textbf{NSG\_uni}	  & \underline{0.477}  & \textbf{0.575}      & \textbf{0.500}   & \underline{0.424}     & \underline{0.324}  & \underline{0.515}      & \underline{0.359}   & \underline{0.229}  \\
HAKE+\textbf{{\model}-Adv}	  & \textbf{0.498}  & \textbf{0.584}      & \textbf{0.514}   & \textbf{0.451}     & \textbf{0.337}  & \textbf{0.533}      & \textbf{0.374}   & \textbf{0.239}  \\
\bottomrule
\end{tabular}}
\caption{Link prediction results on RotatE and HAKE. \textbf{{\model}-Adv} denotes {\model} based on self-adversarial sampling, \textbf{Bold} numbers are the best results for each type of model. Underlined numbers mean the best performances of baselines.}
\label{reulst2}
\end{table*}

%% file: Tables/experiment3.tex
\begin{table*}[t]
\centering
\setlength{\tabcolsep}{4.0mm}
{
\footnotesize
\begin{tabular}{l|cc|cc}
\toprule
\multirow{2}*{Semantic Matching Models} & \multicolumn{2}{c|}{WN18RR} & \multicolumn{2}{c}{FB15K237} \\
	& MRR	& Hits@10	& MRR	& Hits@10	\\

\hline
DistMult+Uniform          &  0.412    &  0.463 & 0.213    & 0.383    \\
DistMult+Bernoulli          &  0.396   &  0.437 & 0.262    & 0.430   \\
DistMult+NSCaching       &0.413     &0.455                      &0.288 & 0.458     \\
DistMult+Self-Adv &0.416 &0.463 & 0.215   & 0.395  \\
% --------addtional information
DistMult+ESNS &\underline{0.424}  &\underline{0.488}  &\underline{0.296}  &\underline{0.465}            \\

DistMult+\textbf{{\model}-Adv}	   &\textbf{0.439}  &\textbf{0.535}   & \textbf{0.301}  & \textbf{0.470}       \\
\midrule
ComplEx+Uniform          &  0.429    &  0.478& 0.214    & 0.387    \\
ComplEx+Bernoulli          &  0.405    &  0.441 & 0.268    & 0.442  \\
ComplEx+NSCaching       &0.446     &0.509                       &0.302 & \underline{\textbf{0.481}}     \\
ComplEx+Self-Adv &0.435 &0.493 & 0.211    & 0.395    \\
% --------addtional information
ComplEx+ESNS &\underline{0.450}  &\underline{0.512}  &\underline{0.303}  &0.471            \\

ComplEx+\textbf{{\model}-Adv}	   &\textbf{0.468}  &\textbf{0.552}   & \textbf{0.307}  & 0.479       \\

\bottomrule
\end{tabular}}
\caption{Link prediction results for semantic matching KGE models on two datasets. \textbf{{\model}-Adv} denotes {\model} based on self-adversarial sampling, \textbf{Bold} numbers are the best results for each type of model. Underlined numbers mean the best performances of baselines. All baseline results are from ESNS\cite{esns}.}
\label{result3}
\end{table*}

%% file: Tables/combine.tex
\begin{table*}[t]
\setlength{\tabcolsep}{4.1mm}
\centering
\begin{tabular}{c|cccc}
\toprule
% \multirow{2}*{Models} & \multicolumn{4}{c}{FB15K237} \\
	& MRR	& Hits@10	& Hits@3  & Hits@1 \\
\midrule
HAKE+Uniform &0.304 &0.482 &0.333 &0.216	\\
HAKE+{\model}-Uni &\textbf{0.332} &\textbf{0.524} &\textbf{0.368 }&\textbf{0.236}   \\
\midrule
HAKE+RW-SANS  &0.305  &0.488   &0.336   &0.214  \\
HAKE+{\model}-RW-SANS & \textbf{0.322}     &  \textbf{ 0.515 }    & \textbf{
0.353 }    & \textbf{0.228}  \\
\bottomrule
\end{tabular}
\caption{{\model} upon different negative sampling methods on FB15K237 with HAKE as KGE.}
\label{tab:combine}
\end{table*}

%% file: Tables/ablation.tex
\begin{table*}[t]
\setlength{\tabcolsep}{6.7mm}
\centering
\begin{tabular}{c|cccc}
\toprule
% \multirow{2}*{Models} & \multicolumn{4}{c}{FB15K237} \\
Models	& MRR	& Hits@10	& Hits@3  & Hits@1 \\
\midrule
{\model}	                          & \textbf{0.337}  & 0.533      & \textbf{0.374}   & \textbf{0.239}	\\
-WARM                            	 &0.336      &    \textbf{0.534}   & \textbf{0.374}     & 0.237  \\
-MPNE                             	  &  0.306    &   0.509     &  0.341    &0.207   \\
-AdaMix                                   & 0.288     &0.471        &   0.318   & 0.198  \\
\bottomrule
\end{tabular}
\caption{Ablation study of {\model} on FB15K237 with HAKE as KGE.}
\label{tab:ablation}
% \vspace{-0.5cm}
\end{table*}

%% file: Sections/2_Related_Work.tex
\section{Related Work}

\subsubsection{KGE models}
The general learning approach of KGEs is to define a score function to measure the plausibility of triples. Traditional KGE Methods consist of translational distance-based models and semantic matching models. Translational distance-based models such as TransE \cite{TransE} and SE\cite{se} calculate the Euclidean distance between the relational projection of entities. Semantic matching models such as DistMult \cite{DistMult} and ComplEx \cite{ComplEx} use similarity-based scoring function to measure the correctness of triples. The recent KGE approach attempt to model some of the properties present in KGs, such as RotatE~\cite{RotatE} aims to model and infer various relation patterns by projecting entities into complex space, HAKE~\cite{hake} aims to model semantic hierarchies in KGs by mapping entities into the polar coordinate system, and COMPGCN ~\cite{compgcn} uses deep neural networks to embed KGs. KGE approaches can be effectively applied to the task of knowledge graph complementation, also known as the link prediction task. However, the ability of KGE models to discriminate whether a triple is correct is often affected by the quality of the negative triples used in the training phase.

% According to the score function,
% KGE models can be divided into two main categories. One score function type is distance-based functions which measure the plausibility of facts by calculating the distance between entities, including the intuitive distance-based
% approach by calculation the Euclidean distance between the relational projection of entities, such as SE \cite{se} and some models treating relations as translations from head entities to tail entities, such as TransE \cite{TransE}, TransH \cite{TransH}, TransR \cite{TransR}. Another type of score function is similarity-based functions which follow the premise of semantic matching to measure the rationality of the triples, including ComplEx \cite{ComplEx}, DistMult \cite{DistMult}, TuckER \cite{tucker}.
% KGE approaches can be effectively applied to the task of knowledge graph complementation, also known as the link prediction task. However, the ability of KGE models to discriminate whether a triple is correct is often affected by the quality of the negative triples used in the training phase.

\subsubsection{Negative Sampling for KGE}
% Negative sampling techniques for KGE generally follow the local closed-world assumption by taking a portion of samples from triples that are not observed in KG as negative triples and allowing the model to learn to distinguish the observed positive triples among these negative triples. 
The existing negative sampling techniques for KGE can be classified into two categories: 1) 
Negative sampling methods that ignore false-negative triples. This kind of approach is based on the closed-world assumption, where all unlabeled corrupted triples are considered as negative triples, and search
for high-quality negative triples. Uniform sampling\cite{TransE} method samples negative triples from a uniform distribution. KBGAN \cite{KBGAN} and IGAN \cite{igan} use the generative adversarial framework to feed the model with high-quality negative triples.
Self-adversarial sampling \cite{RotatE} performs similarly to KBGAN, but it gives different training weights to negative triples. To achieve effectiveness and efficiency, NSCaching \cite{zhang2019nscaching} maintains a cache containing candidates of negative triples.
2) Negative sampling considering false-negative triples. This type of method
aims to minimize the chance of sampling
false-negative triples, such as Bernoulli sampling~\cite{TransH}, ESNS~\cite{esns} and CANS~\cite{niu2022cake}. For example, CANS leverages external commonsense information and the characteristics of complex relations to model false-negative triples’ distribution, then give lower training weights to false-negative triples following the self-adversarial negative sampling loss \cite{RotatE}. However, all previous negative sampling methods are searching-based methods, which inefficiently search for high-quality negative triples from a large set of unlabeled corrupted triples. Besides, CANS requires costly manual effort to gather valuable external information, and Bernoulli sampling does not use external information, but this method is a fixed sampling scheme. Our approach differs from these methods in that we do not avoid sampling false-negative triples, but rather refine negative triples to high-quality triples as much as possible.

\subsubsection{Mixup Method}
Mixup \cite{MixGCF2021} is a data augmentation method that generates a new instance by convex combinations of pairs of training instances. Despite its simplicity, it has shown that it can improve the generalization and Robustness of the model in many applications \cite{conf/nips/ThulasidasanCBB19,journals/corr/abs-2006-06049}. Further, a number of modified mixup versions have been proposed for supervised and unsupervised learning. For supervised learning, Mixup \cite{MixGCF2021} for the first time linearly interpolates two samples and their corresponding labels to generate virtual samples, and experimental results show that mixup is generally applicable to image, speech, and table datasets. In unsupervised scenarios, mixup method is mainly used to construct high-quality virtual negative samples to effectively improve the generalization and robustness of models. MixGCF \cite{MixGCF2021} integrates multiple negative samples to synthesize a difficult negative sample by positive mixing and hop mixing to improve the performance of GNN-based recommendation models. More similar to us are methods that leverage off-the-shelf mixup methods to generate harder negative triples for KGE such as MixKG \cite{mixkg}. Different from existing methods, {\model} design an adaptive mixup mechanism to dynamically refine noisy negative triples.

%% file: Sections/6_Conclusion.tex
\section{Conclusion and Future Work}
In this paper, we explore the denoising issue in sampling high-quality negative triples for KGEs and find these noisy triples have a significant impact on the performance of KGE. Further, we propose a novel and easily pluggable method to alleviate the denoising issue in negative sampling for KGEs. Empirical results on two benchmark datasets demonstrate the effectiveness of our approach. In the future, we plan to extend to recognize noisy triples with unseen patterns in the training set and apply active learning to our method.

\paragraph*{Supplemental Material Statement:} Source code and datasets are available for reproducing the results.\footnote{https://github.com/DeMix2023/Demix}

%% file: main_iswc.bbl
\begin{thebibliography}{10}
\providecommand{\url}[1]{\texttt{#1}}
\providecommand{\urlprefix}{URL }
\providecommand{\doi}[1]{https://doi.org/#1}

\bibitem{sans}
Ahrabian, K., Feizi, A., Salehi, Y., Hamilton, W.L., Bose, A.J.: Structure
  aware negative sampling in knowledge graphs. In: Proceedings of the
  Conference on Empirical Methods in Natural Language Processing. pp.
  6093--6101 (2020)

\bibitem{mixmatch}
Berthelot, D., Carlini, N., Goodfellow, I., Papernot, N., Oliver, A., Raffel,
  C.A.: Mixmatch: A holistic approach to semi-supervised learning. In: Advances
  in Neural Information Processing Systems. vol.~32 (2019)

\bibitem{freebase}
Bollacker, K., Evans, C., Paritosh, P., Sturge, T., Taylor, J.: Freebase: A
  collaboratively created graph database for structuring human knowledge. In:
  Proceedings of the ACM Conference on Management of Data. pp. 1247--1250
  (2008)

\bibitem{TransE}
Bordes, A., Usunier, N., Garcia-Duran, A., Weston, J., Yakhnenko, O.:
  Translating embeddings for modeling multi-relational data. In: Proceedings of
  the Annual Conference on Neural Information Processing Systems. pp.
  2787--2795 (2013)

\bibitem{se}
Bordes, A., Weston, J., Collobert, R., Bengio, Y.: Learning structured
  embeddings of knowledge bases. In: Proceedings of the AAAI Conference on
  Artificial Intelligence. p. 301–306 (2011)

\bibitem{KBGAN}
Cai, L., Wang, W.Y.: {KBGAN:} adversarial learning for knowledge graph
  embeddings. In: {NAACL-HLT}. pp. 1470--1480 (2018)

\bibitem{journals/corr/abs-2006-06049}
Carratino, L., Ciss{\'e}, M., Jenatton, R., Vert, J.P.: On mixup
  regularization. arXiv preprint arXiv:2006.06049  (2020)

\bibitem{mixkg}
Che, F., Yang, G., Shao, P., Zhang, D., Tao, J.: Mixkg: Mixing for harder
  negative samples in knowledge graph. arXiv preprint arXiv:2202.09606  (2022)

\bibitem{incremental}
Chen, T.S., Hung, W.C., Tseng, H.Y., Chien, S.Y., Yang, M.H.: Incremental false
  negative detection for contrastive learning (2021)

\bibitem{dettmers2018conve}
Dettmers, T., Pasquale, M., Pontus, S., Riedel, S.: Convolutional 2d knowledge
  graph embeddings. In: Proceedings of the AAAI Conference on Artificial
  Intelligence. pp. 1811--1818 (2018)

\bibitem{gan}
Goodfellow, I., Pouget-Abadie, J., Mirza, M., Xu, B., Warde-Farley, D., Ozair,
  S., Courville, A., Bengio, Y.: Generative adversarial nets. In: Advances in
  Neural Information Processing Systems. pp. 2672--2680 (2014)

\bibitem{qa}
Hao, Y., Zhang, Y., Liu, K., He, S., Liu, Z., Wu, H., Zhao, J.: An end-to-end
  model for question answering over knowledge base with cross-attention
  combining global knowledge. In: Proceedings of Annual Meeting of the
  Association for Computational Linguistics. pp. 221--231 (2017)

\bibitem{MixGCF2021}
Huang, T., Dong, Y., Ding, M., Yang, Z., Feng, W., Wang, X., Tang, J.: Mixgcf:
  An improved training method for graph neural network-based recommender
  systems. In: Proceedings of the ACM Knowledge Discovery and Data Mining
  (2021)

\bibitem{Ji_2022}
Ji, S., Pan, S., Cambria, E., Marttinen, P., Yu, P.S.: A survey on knowledge
  graphs: Representation, acquisition, and applications. {IEEE} Transactions on
  Neural Networks and Learning Systems  \textbf{33}(2),  494--514 (2022)

\bibitem{adam}
Kingma, D.P., Ba, J.: Adam: {A} method for stochastic optimization. In:
  Proceedings of the International Conference on Learning Representations
  (2015)

\bibitem{niu2022cake}
Niu, G., Li, B., Zhang, Y., Pu, S.: Cake: A scalable commonsense-aware
  framework for multi-view knowledge graph completion. In: Proceedings of the
  Annual Meeting of the Association for Computational Linguistics (2022)

\bibitem{pytorch}
Paszke, A., Gross, S., Massa, F., Lerer, A., Bradbury, J., Chanan, G., Killeen,
  T., Lin, Z., Gimelshein, N., Antiga, L., Desmaison, A., K{\"{o}}pf, A., Yang,
  E.Z., DeVito, Z., Raison, M., Tejani, A., Chilamkurthy, S., Steiner, B.,
  Fang, L., Bai, J., Chintala, S.: Pytorch: An imperative style,
  high-performance deep learning library. In: Proceedings of the Annual
  Conference on Neural Information Processing Systems. pp. 8024--8035 (2019)

\bibitem{RotatE}
Sun, Z., Deng, Z., Nie, J., Tang, J.: Rotate: Knowledge graph embedding by
  relational rotation in complex space. In: Proceedings of the International
  Conference on Learning Representations (2019)

\bibitem{conf/nips/ThulasidasanCBB19}
Thulasidasan, S., Chennupati, G., Bilmes, J.A., Bhattacharya, T., Michalak, S.:
  On mixup training: Improved calibration and predictive uncertainty for deep
  neural networks. In: Proceedings of the Annual Conference on Neural
  Information Processing Systems. pp. 13888--13899 (2019)

\bibitem{fb15k237}
Toutanova, K., Chen, D., Pantel, P., Poon, H., Choudhury, P., Gamon, M.:
  Representing text for joint embedding of text and knowledge bases. In:
  Proceedings of the Conference on Empirical Methods in Natural Language
  Processing. pp. 1499--1509 (2015)

\bibitem{ComplEx}
Trouillon, T., Welbl, J., Riedel, S., Gaussier, {\'E}., Bouchard, G.: Complex
  embeddings for simple link prediction. In: Proceedings of the International
  Conference on Machine Learning. pp. 2071--2080 (2016)

\bibitem{compgcn}
Vashishth, S., Sanyal, S., Nitin, V., Talukdar, P.: Composition-based
  multi-relational graph convolutional networks. In: International Conference
  on Learning Representations (2020)

\bibitem{igan}
Wang, P., Li, S., Pan, R.: Incorporating {GAN} for negative sampling in
  knowledge representation learning. In: Proceedings of the Thirty-Second
  {AAAI} Conference on Artificial Intelligence. pp. 2005--2012 (2018)

\bibitem{TransH}
Wang, Z., Zhang, J., Feng, J., Chen, Z.: Knowledge graph embedding by
  translating on hyperplanes. In: Proceedings of the Twenty-Eighth {AAAI}
  Conference. pp. 1112--1119 (2014)

\bibitem{long-tail}
Wang, Z., Lai, K., Li, P., Bing, L., Lam, W.: Tackling long-tailed relations
  and uncommon entities in knowledge graph completion. In: Proceedings of the
  2019 Conference on Empirical Methods in Natural Language Processing and the
  9th International Joint Conference on Natural Language Processing
  (EMNLP-IJCNLP). pp. 250--260 (2019)

\bibitem{xiong-2017-explicit}
Xiong, C., Power, R., Callan, J.: Explicit semantic ranking for academic search
  via knowledge graph embedding. In: Proceedings of the International World
  Wide Web Conferences. pp. 1271--1279 (2017)

\bibitem{DistMult}
Yang, B., Yih, W., He, X., Gao, J., Deng, L.: Embedding entities and relations
  for learning and inference in knowledge bases. In: Proceedings of the
  International Conference on Learning Representations (2015)

\bibitem{graphdialog}
Yang, S., Zhang, R., Erfani, S.: {G}raph{D}ialog: Integrating graph knowledge
  into end-to-end task-oriented dialogue systems. In: Proceedings of the 2020
  Conference on Empirical Methods in Natural Language Processing (EMNLP). pp.
  1878--1888 (2020)

\bibitem{esns}
Yao, N., Liu, Q., Li, X., Yang, Y., Bai, Q.: Entity similarity-based negative
  sampling for knowledge graph embedding. In: Proceedings of the 19th Pacific
  Rim International Conference on Artificial Intelligence, {PRICAI}. pp. 73--87
  (2022)

\bibitem{XTransE}
Zhang, W., Deng, S., Wang, H., Chen, Q., Zhang, W., Chen, H.: Xtranse:
  Explainable knowledge graph embedding for link prediction with lifestyles in
  e-commerce. In: Proceedings of the Joint International Semantic Technology
  Conference. vol.~1157, pp. 78--87 (2019)

\bibitem{zhang2019nscaching}
Zhang, Y., Yao, Q., Shao, Y., Chen, L.: Nscaching: Simple and efficient
  negative sampling for knowledge graph embedding. In: Proceedings of the IEEE
  International Conference on Data Engineering. pp. 614--625 (2019)

\bibitem{hake}
Zhang, Z., Cai, J., Zhang, Y., Wang, J.: Learning hierarchy-aware knowledge
  graph embeddings for link prediction. In: Thirty-Fourth {AAAI} Conference on
  Artificial Intelligence. pp. 3065--3072 (2020)

\end{thebibliography}
